# Confounder Analysis in Measuring Representation in Product Funnels

Jilei Yang
jlyang@linkedin.com
LinkedIn Corporation

Wentao Su
wesu@linkedin.com
LinkedIn Corporation

## Abstract

This paper discusses an application of Shapley values in the causal inference field, specifically on how to select the top confounder variables for coarsened exact matching method in a scalable way. We use a dataset from an observational experiment involving LinkedIn members as a use case to test its applicability, and show that Shapley values are highly informational and can be leveraged for its robust importance-ranking capability.

## Introduction

As web and social media companies are under increasing scrutiny about fairness issues related to their products, researchers and developers are focusing more on designing new methods that offer user inclusivity and transparency. In this context, one significant effort is a new methodology to measure the representation in an equity framework through funnel analysis, where in each product funnel stage it measures how the equity representation such as gender representation performs. The detailed methodology for funnel representation measurement is discussed in [1], where the Coarsened Exact Matching (CEM) algorithm [2] from causal inference field [12] is used to adjust the effects of confounders in estimating the new metric designed, called "adjusted funnel survival ratio". This new metric measures the gender funnel survival ratio (gender ratio in the current funnel / gender ratio in the previous funnel) after considering the confounding effect that is external to our own platform, in order to help us track the equity progress consistently across different products at LinkedIn.

One question that remains to be answered is, among the confounders used in our funnel analysis, which one plays a more important role in contributing to the confounding effect. Businesses usually have existing repositories of confounders such as member profile features (e.g., country, industry). It is also important to learn the impact of each confounder on the funnel survival ratio directly from the data, i.e., the contribution of each confounder in adjusting the funnel survival ratio after considering its effect. By learning the contribution of each confounder, we can have a priority list of confounders for deep-dive analysis (e.g., demographic disparity of each confounder). And from the perspective of implementation, we can conduct confounder selection to filter out confounders that are irrelevant to funnel survival ratio estimation when the number of confounders is large, making the implementation of the method more efficient. This paper discusses the application of Shapley value methodology to rank order the importance of confounding variables and generate insights on how each confounder impacts the core funnel metric.

# Proposed Approach

## Shapley value

The following concepts from coalitional game theory are used in the formalization of our proposed approach. A coalitional game is a tuple $(N, v)$, where $N = \{1, 2, ..., n\}$ is a set of $n$ players, and $v: 2^n \rightarrow R$ is a characteristic function such that $v(\emptyset) = 0$. Subsets of $N$ are coalitions and $N$ is referred to as the grand coalition of all players, and the function $v$ describes the value (reward) of each coalition. The goal is to split the value of the grand coalition $v(N)$ among all players in a "fair" way. The solution is an operator ϕ which assigns to $(N, v)$ a vector $\phi(v) = (\phi_1(v), ..., \phi_n(v)) \in R^n$. For each game with at least one player there are infinitely many solutions, some of which are more "fair" than others. The following four statements are attempts at axiomatizing the notion of "fairness" of a solution ϕ.

- Efficiency Axiom: $\sum_{i \in N} \phi_i(v) = v(N)$.
- Symmetry Axiom: If for two players $i$ and $j$, $v(S \cup \{i\}) = v(S \cup \{j\})$ holds for every $S \subset N$ where $i, j \notin S$, then $\phi_i(v) = \phi_j(v)$.
- Dummy Axiom: If $v(S \cup \{i\}) = v(S)$ holds for every $S \subset N$ where $i \notin S$, then $\phi_i(v) = 0$.
- Additivity Axiom: For any pair of games $v, w$: $\phi(v + w) = \phi(v) + \phi(w)$, where $(v + w)(S) = v(S) + w(S)$ for all $S \subset N$.

Theorem: For the game $(N, v)$ there exists a unique solution ϕ, which satisfies axioms 1 to 4 and it is the **Shapley value** [3]:

$$\phi_i(v) = \sum_{S \subset N \setminus \{i\}} \frac{|S|!(n-|S|-1)!}{n!} (v(S \cup \{i\}) - v(S)), i = 1, ..., n.$$

The Shapley value has a wide range of applications. For example, recently it has been widely used in model interpretation [4, 5, 6], where popular model interpretation approaches such as KernelSHAP [6], TreeSHAP [7] and cohort Shapley [8, 9] are all based on the Shapley value.

## Shapley Sampling value

The computational complexity of the Shapley value increases exponentially as the number of players $n$ increases. In practice, for large value of $n$ (e.g., $n > 6$), we can use Shapley Sampling value [10, 4, 5] to approximate the Shapley value, where the approximation algorithm can be run in polynomial time.

Let π(N) be the set of all ordered permutations of $N$ (e.g., for $N = \{1, 2, 3\}$, π(N) = $\{(1, 2, 3), (1, 3, 2), (2, 1, 3), (2, 3, 1), (3, 1, 2), (3, 2, 1)\}$). Let $Pre_i(P)$ be the set of players who are predecessors of play $i$ in the ordered permutation $P \in π(N)$ (e.g., for $P = (1, 2, 3)$, $Pre_1(P) = \emptyset$, $Pre_2(P) = \{1\}$, $Pre_3(P) = \{1, 2\}$). It is easy to see that, the Shapley value can now be expressed as:

$$\phi_i(v) = \frac{1}{n!} \sum_{P \in π(N)} [v(Pre_i(P) \cup \{i\}) - v(Pre_i(P))], \; i = 1,...,n.$$

The Shapley Sampling value is based on the above expression, where $m$ samples of ordered permutations are randomly sampled from π(N). The detailed algorithm is summarized in Algorithm 1:

---

Algorithm 1: Approximate the Shapley value of the $i$th player $\phi_i(v)$

---

determine $m$, the desired number of samples
initialize $\phi_i(v) \leftarrow 0$
for 1 to $m$ do
    choose a random permutation $P \in π(N)$
    calculate $v_1 \leftarrow v(Pre_i(P) \cup \{i\})$
    calculate $v_2 \leftarrow v(Pre_i(P))$
    update $\phi_i(v) \leftarrow \phi_i(v) + (v_2 - v_1)$
end for
$\phi_i(v) \leftarrow \phi_i(v)/m$

---

Assume the time complexity of evaluating $v$ on an arbitrary subset of $N$ is $O(V)$, then the time complexity of calculating the exact Shapley value is $O(2^n V)$, and the time complexity of calculating the Shapley Sampling value is reduced to $O(mnV)$. The efficiency of this approximation is discussed in detail in [10]. In short, the central limit theorem is used to determine the sample size $m$, if we want to guarantee that the approximation error is lower than a predefined value $\epsilon$ with a probability greater than a predefined value $1 - α$.

## Application to Confounder Analysis

Assume $N = \{1, 2, ..., n\}$ is a set of $n$ confounders. The adjusted funnel survival ratio can be regarded as a function $r: 2^n \rightarrow R^+$, where $r(\emptyset)$ is the original funnel survival ratio without considering any confounders, $r(N)$ is the adjusted funnel survival ratio in consideration of all $n$ confounders, and $r(S)$ is the adjusted funnel survival ratio in consideration of confounders in $S \subset N$. Following the definition of Shapley value in the above section, we can define the "contribution" of confounder $i$ to the funnel survival ratio adjustment as the Shapley value $\phi_i$:

$$\phi_i = \sum_{S \subset N\setminus\{i\}} \frac{|S|!(n-|S|-1)!}{n!} (r(S \cup \{i\}) - r(S)), \ i = 1, ..., n.$$

According to the Efficiency Axiom of Shapley value, we have $\sum_{i \in N} \phi_i = r(N) - r(\emptyset)$, i.e., the contributions of all $n$ confounders sum up to the difference between the adjusted funnel survival ratio and the original funnel survival ratio.

The Shapley values of confounders can be used in confounder selection. When there are many confounders being used in CEM algorithm, it is likely that in many strata created by CEM, only treatment units or control units are retained where treatment and control refer to the gender category group in equity context. These data points are not used in calculating the adjustment funnel survival ratio, making the calculation of survival ratio less accurate.

Shapley value can help find out a subset of confounders which leads to higher matching rate for treatment and control groups while maintaining a high level of adjusted funnel survival ratio. A straightforward approach is to select the top $k$ confounders with the highest Shapley values, since the most influential confounder variables are well studied in the vertical product team and the top k (k is between 5 and 10) should be good enough to capture most of the confounding effect.

## Comparison with Other Approaches

There exist approaches other than Shapley value which can also be used in determining the contribution of each confounder. Here we list two of them and compare them with Shapley value.

### Add-One value

Add-One value of confounder $i$ is calculated by taking the difference between the adjusted funnel survival ratio in consideration of confounder $i$ only and the original funnel survival ratio without considering any confounders. Following the previous notations, we have the Add-One value of confounder $i$:

$$\phi_{add-one, i} = r(\{i\}) - r(\emptyset).$$

As we can see, the Add-One value of confounder $i$ is one "special" case of Shapley value by replacing $S \subset N\setminus\{i\}$ with $S = \emptyset$ in the summation term. Obviously, it doesn't take into consideration of all possible scenarios on how confounder $i$ contributes to an arbitrary subset of other confounders (as what Shapley value does). As a result, it doesn't possess the desired properties (four statements) as Shapley value does.

### Counterexample

It is easy to see that Add-One value coincides with Shapley value when the adjusted funnel survival ratio $r$ is a linear function of all $n$ confounders, i.e., $r(N) = \sum_{i=1}^{n} r(\{i\})$. In this case, all the confounders are independently contributing to $r$ without any interactions, which actually, rarely happens in real life. A counterexample of using Add-One value instead of Shapley value is presented in the following.

Assume there are only two confounders 1 and 2, and $r$ takes value 1 if and only if both confounders 1 and 2 are present, and takes value 0 otherwise. If we use $x_i$ to denote the presence/absence of confounder $i$, i.e., $x_i = 1$ if confounder $i$ is present, and $x_i = 0$ if confounder $i$ is absent, then $r(x_1, x_2) = x_1 \wedge x_2$. The Add-One values of confounders 1 and 2 are:

$$\phi_{add-one,\ 1} = r(1, 0) - r(0, 0) = 0 - 0 = 0,$$
$$\phi_{add-one,\ 2} = r(0, 1) - r(0, 0) = 0 - 0 = 0.$$

It seems unreasonable to assign zero contributions to both confounders 1 and 2, as they do make contributions to $r$ once they are in a cooperative mode. The Shapley values of confounders 1 and 2 are:

$$\phi_1 = \frac{1}{2}(r(1, 0) - r(0, 0)) + \frac{1}{2}(r(1, 1) - r(0, 1)) = 0 + \frac{1}{2} = \frac{1}{2},$$
$$\phi_2 = \frac{1}{2}(r(0, 1) - r(0, 0)) + \frac{1}{2}(r(1, 1) - r(1, 0)) = 0 + \frac{1}{2} = \frac{1}{2}.$$

The contributions of confounders 1 and 2 assigned by Shapley values seem much more reasonable than those assigned by Add-One values.

### Remove-One value

Remove-One value of confounder $i$ is calculated by taking the difference between the adjusted funnel survival ratio in consideration of all $n$ confounders and the adjusted funnel survival ratio in consideration of all confounders except confounder $i$. Similarly, following the previous notations, we have the Remove-One value of confounder $i$:

$$\phi_{remove-one,\ i} = r(N) - r(N\backslash\{i\}).$$

As we can see, the Remove-One value of confounder $i$ is another "special" case of Shapley value by replacing $S \subset N\backslash\{i\}$ with $S = N\backslash\{i\}$ in the summation term. Similar to Add-One value, it doesn't take into consideration of all possible scenarios on how confounder $i$ contributes to an arbitrary subset of other confounders (as what Shapley value does). As a result, it doesn't possess the desired properties (four statements) as Shapley value does.

### Counterexample

Similar to Add-One value, Remove-One value also coincides with Shapley value when the adjusted funnel survival ratio $r$ is a linear function of all $n$ confounders, i.e., $r(N) = \sum_{i=1}^{n} r(\{i\})$, which rarely happens in real life. A counterexample of using Remove-One value instead of Shapley value is presented in the following.

Assume there are only two confounders 1 and 2, and $r$ takes value 1 if and only if at least one of two confounders 1 and 2 are present, and takes value 0 otherwise. If we use $x_i$ to denote the presence/absence of confounder $i$, similar to the previous counterexample, then $r(x_1, x_2) = x_1 \vee x_2$. The Remove-One values of confounders 1 and 2 are:

$$\phi_{remove-one, 1} = r(1, 1) - r(0, 1) = 1 - 1 = 0,$$
$$\phi_{remove-one, 2} = r(1, 1) - r(1, 0) = 1 - 1 = 0.$$

It seems unreasonable to assign zero contributions to both confounders 1 and 2, as they do make contributions to $r$ either in an individual mode or in a cooperative mode. The Shapley values of confounders 1 and 2 are:

$$\phi_1 = \tfrac{1}{2}(r(1,0) - r(0,0)) + \tfrac{1}{2}(r(1,1) - r(0,1)) = \tfrac{1}{2} + 0 = \tfrac{1}{2},$$
$$\phi_2 = \tfrac{1}{2}(r(0,1) - r(0,0)) + \tfrac{1}{2}(r(1,1) - r(1,0)) = \tfrac{1}{2} + 0 = \tfrac{1}{2}.$$

The contributions of confounders 1 and 2 assigned by Shapley values seem much more reasonable than those assigned by Remove-One values.

## Implementation Detail

Current implementation uses Shapley sampling value to approximate the Shapley value based on Algorithm 1. When the number of confounder variables is less than or equal to 6, the original formula is leveraged to derive the Shapley value. Otherwise, we would randomly draw 100 permutations of confounders to compute the Shapley value.

The prototype script is working on the onboarding completion funnel use case where we measure the survival ratios for newly registered members who completed the full onboarding process. Currently there are 8 confounding variables. Now we can use the Shapley value method to explain which confounder explains the most for the difference between adjusted survival ratio and raw survival ratio. This difference is considered as the confounding effect.

# Confounder Ranking Result

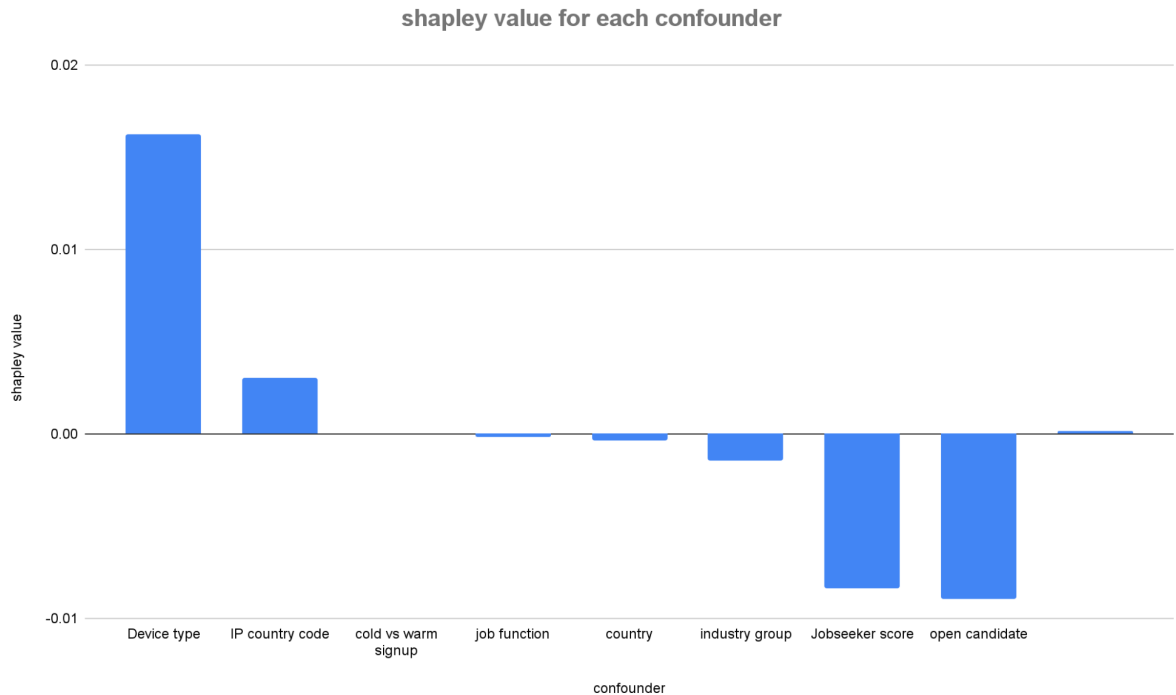

Figure 1. Shapley value for each confounder in onboarding completion funnel use case.

Figure 1 shows the Shapley value for each confounder in the onboarding completion funnel use case. Based on the ranking order we find that device type turns out to be the most important factor that contributes to the difference between raw survival ratio and adjusted survival ratio, meaning that this external factor has strong gender skewness that benefits males more than females which affects the funnel gender representation. To our surprise, country and job function - two most important member profile variables - have nearly zero Shapley value. It may indicate that members from different countries or job functions did not have a significant gender gap of onboarding experience. Another interesting finding is that the job seeker score and open candidate dummy variables show big negative values, meaning that these two confounders are important but the gender skewness benefits females before the adjustment. Based on the result, users should use at least four confounders in funnel analysis: Device type, IP country code, job seeker score and open candidate variables.

In addition, dividing each confounder's Shapley value by the sum of all confounders' would give us the percentage of contribution of each confounder. We also verify that the sum of the Shapley value for 8 variables is equal to the difference between two survival ratios when all 8 confounders are included. This is an illustration of how we can leverage Shapley value to screen out the most important confounders. In reality, all 8 confounders can be supported by the existing representation analysis tool, but with more confounding variables, the importance ranking becomes more useful to the users to expedite the analysis.

# Discussion

This paper discusses the use of Shapley value in causal inference field, specifically on how to select the top confounder variables for coarsened exact matching methods in a scalable way. This research can be extended to confounder selection process in other causal inference methodologies, and multi-touch marketing attribution in which case the credit of conversion is assigned to each unique path of the marketing channel. There are already some initial outcomes [11] in these areas but more research is needed to expand the scope of Shapley value applications.

# Acknowledgements

We would like to express our sincere thanks to YinYin Yu and Humberto Gonzalez for their helpful comments and feedback. We also thank our management team Parvez Ahammad and Rahul Tandra for their continuous support.

# Appendix

## Pseudo-code for Implementation

*Loop each confounder value from confounderList*
*confounderList.par.map { case m =>*
    *Initialize the shapley value $\phi_m(v)$*
    *val permutation = permutations of confounders*

    *while (permutation.hasNext && sum < iterationLimit) {*
        *val confounderWithoutM = all confounders in permutation.next() before m*
        *val confounderWithM = confounderWithoutM + m*

        *calculate the CEM-adjusted survival ratio using the MP function FunnelCompute*
        *val survivalRatioDiffWithoutM = FunnelCompute(..., confounderWithoutM)*
        *val survivalRatioDiffWithM = FunnelCompute(..., confounderWithM)*

        *Calculate the incremental change of the two survival ratios*
        *$\phi_m(v) \leftarrow \phi_m(v)$ + (survivalRatioDiffWithM - survivalRatioDiffWithoutM)*
    *}*
    *each confounder shapley value = $\phi_m(v)/m$*
*}*